\documentclass{article}
\usepackage{spconf,amssymb,amsmath,epsfig}
\usepackage{scalerel}
\usepackage{microtype}
\usepackage{graphicx}
\usepackage{multirow}
\usepackage{booktabs}
\usepackage{amssymb,amsmath,graphicx}
\usepackage{caption}
\usepackage{subcaption}
\usepackage[export]{adjustbox}
\usepackage{moredefs}
\usepackage{color}

\defcommand{\vec}[1]{\mathbf{#1}} 

\def\vx{\mathbf{x}}

\def\vy{\mathbf{y}}

\def\sy{y}
\def\hy{\hat{y}}
\def\hvy{\hat{\vy}}

\def\A{{\cal A}}
\def\Z{{\cal Z}}

\def\sos{{\left<\texttt{sos}\right>}}

\def\blank{{\left<\texttt{b}\right>}}

\title{Streaming End-to-End Speech Recognition for Mobile Devices}
\name{Yanzhang He\textsuperscript{*}, Tara N. Sainath\sthanks{Equal contribution}, Rohit Prabhavalkar, Ian McGraw, Raziel Alvarez, Ding Zhao,}
\nameplus{David Rybach, Anjuli Kannan, Yonghui Wu, Ruoming Pang, Qiao Liang, Deepti Bhatia, Yuan Shangguan,}
\nameplusplus{Bo Li, Golan Pundak, Khe Chai Sim, Tom Bagby, Shuo-yiin Chang, Kanishka Rao, Alexander Gruenstein} 

\address{Google, Inc., USA \\
\fontsize{9}{9}\selectfont\ttfamily\upshape
\{yanzhanghe, tsainath\}@google.com}

\begin{document}
\maketitle
\ninept
\begin{abstract}
End-to-end (E2E) models, which directly predict output character sequences given
input speech, are good candidates for on-device speech recognition.
E2E models, however, present numerous challenges:
In order to be truly useful, such models must decode speech utterances in a
streaming fashion, in real time; they must be robust to the long tail of use cases;
they must be able to leverage user-specific context (e.g., contact lists); and
above all, they must be extremely accurate.
In this work, we describe our efforts at building an E2E speech recognizer using
a recurrent neural network transducer.
In experimental evaluations, we find that the proposed approach can outperform a
conventional CTC-based model in terms of both latency and accuracy in a number
of evaluation categories.
\end{abstract}

\section{Introduction \label{sec:introduction}}
The last decade has seen tremendous advances in automatic speech recognition
(ASR) technologies fueled by research in deep neural
networks~\cite{HintonDengYuDahlEtAl12}.
Coupled with the tremendous growth and adoption of smartphones, tablets and
other consumer devices, these improvements have resulted in speech becoming one
of the primary modes of interaction with such devices~\cite{Cohen08,
SchalkwykBeefermanBeaufaysByrneEtAl10}.
The dominant paradigm for recognizing speech on mobile devices is to stream
audio from the device to the server, while streaming decoded results back to the
user.
Replacing such a server-based system with one that can \emph{run entirely
on-device} has important implications from a reliability, latency, and privacy
perspective, and has become an active area of research.  Prominent examples
include \emph{wakeword detection} (i.e., recognizing specific words or
phrases)~\cite{SainathParada15,ArikKlieglChildHestnessEtAl17,He17},
as well as large vocabulary continuous speech recognition
(LVCSR)~\cite{WaibelBadranBlackFrederkingEtAl03,
McGrawPrabhavalkarAlvarezArenasEtAl16}.

Previous attempts at building on-device LVCSR systems have typically consisted
of shrinking traditional components of the overall system (acoustic (AM),
pronunciation (PM), and language (LM) models) to satisfy computational
and memory constraints.
While this has enabled parity in accuracy for narrow domains such as voice
commands and dictation~\cite{McGrawPrabhavalkarAlvarezArenasEtAl16}, performance
is significantly worse than a large server-based system on challenging
tasks such as voice search.

In contrast to previous approaches, we instead focus on building a streaming
system based on the recent advances in end-to-end (E2E) models~\cite{Chan15,
Bahdanau16, Graves12, KimHoriWatanabe17, ChiuSainathWuPrabhavalkarEtAl18}.
Such models replace the traditional components of an ASR system with a single,
end-to-end trained, all-neural model which directly predicts character
sequences, thus greatly simplifying training and inference.
E2E models are thus extremely attractive for on-device applications.

Early E2E work examined connectionist temporal classification
(CTC)~\cite{GravesFernandezGomezSchmidhuber06} with grapheme or word
targets~\cite{GravesJaitly14,HannunCaseCasperCatanzaroEtAl14,
MiaoGowayyedMetze15,SoltauLiaoSak17}.
More recent work has demonstrated that performance can be improved further using
either the recurrent neural network transducer (RNN-T)
model~\cite{Graves12, GravesMohamedHinton13,
RaoSakPrabhavalkar17} or attention-based encoder-decoder
models~\cite{Chan15, KimHoriWatanabe17,
ChiuSainathWuPrabhavalkarEtAl18,ChiuRaffel17}.
When trained on sufficiently large amounts of acoustic training data
($10,000$+ hours), E2E models can outperform conventional hybrid RNN-HMM
systems~\cite{RaoSakPrabhavalkar17, ChiuSainathWuPrabhavalkarEtAl18}.
Most E2E research has focused on systems which process the full input utterance
before producing a hypothesis; models such as RNN-T~\cite{Graves12,
GravesMohamedHinton13} or streaming attention-based models (e.g.,
MoChA~\cite{ChiuRaffel17}) are suitable if streaming recognition is desired.
Therefore, in this work, we build a streaming E2E recognizer based on the RNN-T
model.

Running an end-to-end model on device in a \emph{production environment}
presents a number of challenges: first, the model needs to be at least as
accurate as a conventional system, without increasing latency (i.e., the delay
between the user speaking and the text appearing on the screen), thus running
at or faster than real-time on mobile devices; second, the model should be able to
leverage on-device user context (e.g., lists of contacts, song names, etc.) to
improve recognition accuracy~\cite{Petar15}; finally, the system must be able to
correctly recognize the `long tail' of possible utterances, which is a challenge
for an E2E system trained to produce text directly in the \emph{written domain}
(e.g., \texttt{call two double four triple six five} $\to$ \texttt{call
244-6665}).

In order to achieve these goals, we explore a number of improvements to the
basic RNN-T model: using layer normalization~\cite{Ba16} to stabilize training;
using large batch size~\cite{Sim17}; using word-piece
targets~\cite{SchusterNakajima12}; using a time-reduction layer to speed up
training and inference; and quantizing network parameters to reduce memory
footprint and speed up computation~\cite{AlvarezPrabhavalkarBakhtin16}.
In order to enable contextualized recognition, we use a shallow-fusion
approach~\cite{WilliamsKannanAleksicRybachEtAl18, Pundak18}
to bias towards user-specific context, which we find is
on-par with conventional models~\cite{McGrawPrabhavalkarAlvarezArenasEtAl16,Petar15}.
Finally, we characterize a fundamental limitation of vanilla E2E models: their
inability to accurately model the normalization of \emph{spoken} numeric
sequences in the correct \emph{written} form when exposed to unseen examples.
We address this issue by training the model on synthetic data generated using a
text-to-speech (TTS) system~\cite{Aaron17}, which improves performance on
numeric sets by 18--36\% relative.
When taken together, these innovations allow us to decode speech twice as fast
as real time on a Google Pixel phone, which improves word error rate (WER) by
more than 20\% relative to a conventional CTC embedded model on voice search and
dictation tasks.

\section{Recurrent Neural Network Transducer}
\label{sec:rnnt}
\begin{figure}
  \centering
  \includegraphics[width=0.7\columnwidth]{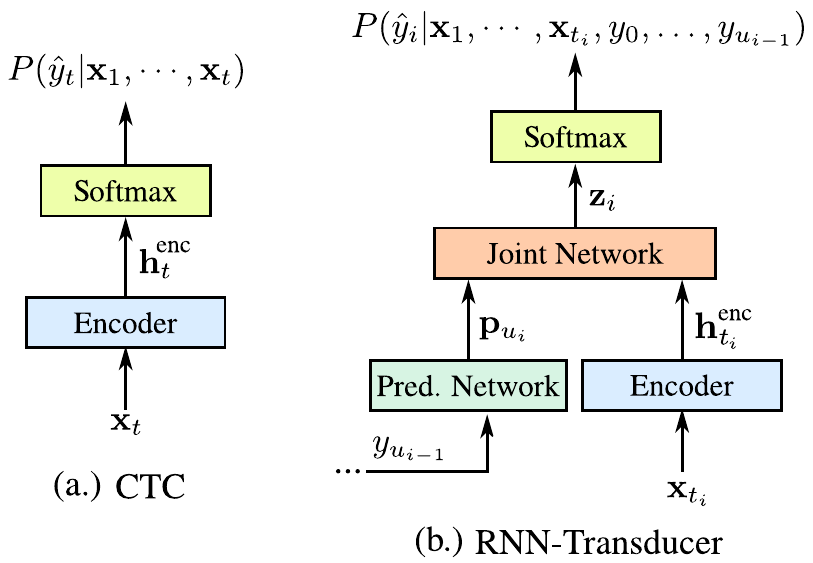}
  \caption{A schematic representation of CTC and RNNT.}
  \label{fig:ctcrnnt-schematic}
\vspace{-0.1in}
\end{figure}
Before describing the RNN-T model in detail, we begin by introducing our
notation.  We denote the parameterized input acoustic frames as $\vx = (\vx_1
\ldots \vx_T)$, where $\vx_t \in \mathbb{R}^d$ are 80-dimensional log-mel
filterbank energies in this work ($d=80$) and $T$ denotes the number of
frames in $\vx$.  We denote the ground-truth label sequence of length $U$ as
$\vy = (\sy_1, \ldots, \sy_{U})$, where $\sy_u \in \Z$ and where $\Z$
corresponds to context-independent (CI) phonemes, graphemes or
word-pieces~\cite{SchusterNakajima12}, in this work.  We sometimes also use a
special symbol, $y_0 = \sos$, which indicates the start of the sentence.

We describe the RNN-T~\cite{Graves12, GravesMohamedHinton13} model by
contrasting it to a CTC~\cite{GravesFernandezGomezSchmidhuber06} model.
CTC computes the distribution of interest, $P(\vy | \vx)$, by augmenting $\Z$
with an additional \emph{blank} symbol, $\blank$, and defining:
\begin{align}
  \small
P(\vy | \vx)&=\hspace{-0.20in} \sum_{\hvy \in \A_{\text{CTC}}(\vx, \vy)} \prod_{t=1}^{T} P(\hy_t | \vx_1, \cdots, \vx_t)
\label{eqn:align-ctc}
\end{align}
\noindent where $\hvy = (\hy_1, \ldots, \hy_T) \in \A_{\text{CTC}}(\vx, \vy)
\subset \{\Z\cup\blank\}^{T}$ correspond to frame-level alignments of length $T$
such that removing blanks and repeated symbols from $\hvy$ yields $\vy$.
CTC makes a strong independence assumption that labels are conditionally
independent of one another given acoustics.
RNN-T removes this independence assumption by instead conditioning on the full
history of previous non-blank labels:
\begin{equation}
  \small
P(\vy | \vx) = \hspace{-0.23in} \sum_{\hy \in \A_{\text{RNNT}}(\vx, \vy)} \prod_{i=1}^{T+U} P(\hy_{i} | \vx_1, \cdots, \vx_{t_i}, y_0, \ldots, y_{u_{i-1}})
\label{eqn:align-rnnt}
\end{equation}
\noindent where $\hat{\vy} = (\hy_1, \ldots, \hy_{T+U}) \in
\A_{\scaleto{RNNT}{3pt}}(\vx, \vy) \subset \{\Z\cup\blank\}^{T+U}$ are alignment
sequences with $T$ blanks and $U$ labels such that removing the blanks in
$\hat{\vy}$ yields $\vy$.  Practically speaking this means that the probability
of observing the $i$th label, $\hat{y}_{i}$, in an alignment, $\hat{\vy}$, is
conditioned on the history of non-blank labels, $y_{1} \ldots y_{u_{i-1}}$,
emitted thus far. Crucially, for both CTC and RNN-T we introduce one final
conditional independence assumption: an alignment label $\hat{y}$ cannot depend
on future acoustic frames. This enables us to build streaming systems that do
not need to wait for the entire utterance to begin processing.

The conditional distributions for both models are parameterized by neural
networks, as illustrated in Figure~\ref{fig:ctcrnnt-schematic}.  Given the input
features we stack unidirectional long short-term memory
(LSTM)~\cite{HochreiterSchmidhuber97} layers to construct an \emph{encoder}.
For CTC the encoder is augmented with a final softmax layer that converts the
encoder output into the relevent conditional probability distribution.
The RNN-T, instead, employs a feed-forward joint network that accepts as input
the results from both the encoder and a \emph{prediction network} that depends
only on label histories.
The gradients required to train both models can be computed using the
forward-backward algorithm
~\cite{Graves12, GravesFernandezGomezSchmidhuber06,GravesMohamedHinton13}.

\section{Real-Time Speech Recognition Using RNN-T}
\label{sec:rt}
This section describes various architectural and optimization
improvements that increase the RNN-T model accuracy and also
allow us to run the model on device faster than real time.

\vspace{-1mm}
\subsection{Model Architecture \label{sec:arch}}
We make a number of architectural design choices for the encoder and prediction
network in RNN-T in order to enable efficient processing on mobile devices.  We
employ an encoder network which consists of eight layers of uni-directional LSTM
cells~\cite{HochreiterSchmidhuber97}.
We add a projection layer~\cite{Hasim14} after each LSTM layer in the encoder,
thus reducing the number of recurrent and output connections.

Motivated by~\cite{Chan15,Soltau2017}, we also add a time-reduction layer in the
encoder to speed up training and inference.
Specifically, if we denote the inputs to the time-reduction layer as
$\mathbf{h}_1, \mathbf{h}_2, \cdots, \mathbf{h}_T$, then we concatenate together
$N$ adjacent input frames to produce $\lceil \frac{T}{N} \rceil$ output frames,
where the $i+1$-th output frame is given by
$[\mathbf{h}_{iN}; \mathbf{h}_{iN+1}; \cdots; \mathbf{h}_{(i+1)N-1}]$,
thus effectively reducing the overall frame rate by a factor of $N$.
The computational savings obtained using a time-reduction layer increase if it
is inserted lower in the encoder LSTM stack.
Applying the time-reduction layer to either model, which already has an
input frame rate of 30ms, has different behaiviors. Specifically, we find that it can be inserted as low as after the
second LSTM layer without any loss in accuracy for RNN-T, whereas
adding it to the CTC phoneme models (with effective output frame rate $\geq$ 60ms)
degrades accuracy.

\vspace{-1mm}
\subsection{Training Optimizations}
In order to stabilize hidden state dynamics of the recurrent layers, we find it
useful to apply layer normalization~\cite{Ba16} to each LSTM layer in the
encoder and the prediction network.
Similar to ~\cite{RaoSakPrabhavalkar17}, we train with word-piece subword
units~\cite{SchusterNakajima12}, which outperform graphemes in our experiments.
We utilize an efficient forward-backward algorithm~\cite{Sim17}, which allows us
to train RNN-T models on tensor processing units (TPUs)~\cite{JouppiEtAl17}.
This allows us to train faster with much larger batch sizes than would be
possible using GPUs, which improves accuracy.

\vspace{-1mm}
\subsection{Efficient Inference}
Finally, we consider a number of runtime optimizations to enable efficient
on-device inference.
First, since the prediction network in RNN-T is analogous to an RNN language
model, its computation is independent of the acoustics.
We, therefore, apply the same state caching techniques used in RNN language
models in order to avoid redundant computation for identical prediction
histories.
In our experiments, this results in saving 50--60\% of the prediction network
computations. In addition, we use different threads for the encoder and the
prediction network to enable pipelining through asynchrony in order to save
time. We further split the encoder execution over two threads corresponding to
the components before and after the time-reduction layer, which balances the
computation between the two encoder components and the prediction network. This
results in a speed-up of 28\% with respect to single-threaded execution.

\vspace{-1mm}
\subsection{Parameter Quantization}
In order to reduce memory consumption, both on disk and at runtime, and to
optimize the model's execution to meet real-time requirements, we quantize
parameters from 32-bit floating-point precision into 8-bit fixed-point, as in our
previous work~\cite{McGrawPrabhavalkarAlvarezArenasEtAl16}.
In contrast to~\cite{McGrawPrabhavalkarAlvarezArenasEtAl16}, we now use a
simpler quantization approach that is linear (as before) but no longer has an
explicit ``zero point'' offset, thus assuming that values are distributed around
floating point zero.
More specifically we define the quantized vector, $\mathbf{x}_q$, to be the
product of the original vector, $\mathbf{x}$, and a quantization factor,
$\theta$, where $\theta = \frac{127}{|\text{max}(\mathbf{x}_\text{min},
\mathbf{x}_\text{max})|}$.
The lack of zero point offset avoids having to apply it prior to performing
operations, such as multiplication, in lower precision thus speeding-up
execution.
Note that we force the quantization to be in the range $\pm (2^7 - 1)$.
Thus, for the typical multiply-accumulate operation, the sum of the products of
2 multiplies is always strictly smaller than 15-bits, which allows us to carry
more than one operation into a 32-bit accumulator, further speeding up inference.
We leverage TensorFlow Lite optimization tools and runtime to execute the model
on both ARM and x86 mobile architectures~\cite{tflite}.
On ARM architectures, this achieves a 3$\times$ speedup compared to floating
point execution.

\vspace{-1.5mm}
\section{Contextual Biasing}
 \label{sec:biasing}
\vspace{-1.5mm}
Contextual biasing is the problem of injecting prior knowledge into an ASR
system during inference, for example a user's favorite songs, contacts, apps or
location~\cite{Petar15}.
Conventional ASR systems perform contextual biasing by building an n-gram finite
state transducer (FST) from a list of biasing phrases, which is composed
on-the-fly with the decoder graph during decoding~\cite{Hall15}.
This helps to bias the recognition result towards the n-grams contained in the
contextual FST, and thus improves accuracy in certain scenarios.
In the E2E RNN-T model, we use a technique similar to~\cite{Hall15}, to compute
biasing scores $P_C(\vec{y})$, which are interpolated with the base model $P(\vec{y}|\vec{x})$ using
shallow-fusion~\cite{Anjuli18} during beam search:
\vspace{-1mm}
\begin{equation}
\vec{y}^* = \underset{\vec{y}}{\arg\max} \log P(\vec{y}|\vec{x}) + \lambda \log P_{C}(\vec{y})
\label{eq:context}
\end{equation}
\noindent
where, $\lambda$ is a tunable hyperparameter controlling how much the contextual
LM influences the overall model score during beam search.

\begin{figure}[h]
  \centering
  \includegraphics[scale=0.4]{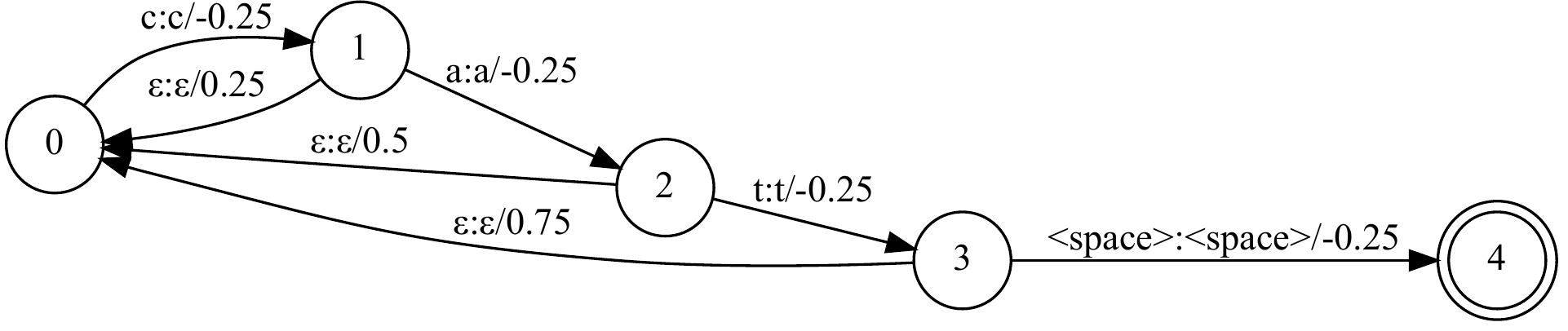}
   \caption{{Contextual FST for the word ``cat'', represented at the subword unit level with backoff arcs.}}
  \label{fig:biasing-fst}
  \vspace{-0.2in}
\end{figure}

To construct the contextual LM, we assume that a set of word-level biasing
phrases are known ahead of time, and compile them into a weighted finite state
transducer (WFST)~\cite{MohriPereiraRiley02}.
This word-level WFST, $G$, is then left-composed with a ``speller'' FST, $S$,
which transduces a sequence of graphemes or word-pieces into the corresponding
word, to obtain the contextual LM: $C = \text{min}(\text{det}(S \circ G))$.
In order to avoid artificially boosting prefixes which match early on but do not
match the entire phrase, we add a special failure arc which \emph{removes} the
boosted score, as illustrated in Figure~\ref{fig:biasing-fst}.
Finally, in order to improve RNN-T performance on proper nouns, which is
critical for biasing, we train with an additional 500M unsupervised voice search utterances
(each training batch is filled with supervised data 80\% of the time and
unsupervised data 20\% of the time).
The unsupervised data is transcribed by our production-level recognizer \cite{Golan16} and filtered to contain high-confidence utterances with proper nouns only. Note that training with this data does not change results on our voice-search and dictation test sets, but only improves performance on the contextual biasing
results described in Table~\ref{table:biasing}.

\vspace{-1mm}
\section{Text Normalization}
 \label{sec:numerics}
\vspace{-1mm}
Conventional models are trained in the \emph{spoken} domain~\cite{Vasserman15,
Sproat17}, which allows them to convert unseen numeric sequences into the
\emph{written} domain during decoding (e.g., \texttt{navigate to two twenty one b baker street}
$\to$ \texttt{navigate to 221b baker street}), which alleviates the data sparsity issue.
This is done by training a
class-based language model where classes such as ADDRESSNUM replace actual
instances in the training data, and training grammar WFSTs that map these
classes to all possible instances through hand-crafted rules.
During decoding, the recognizer first outputs hypotheses in the spoken domain
with the numeric part enclosed in the class tags
(\texttt{<addressnum> two twenty one b </addressnum>}),
which is then converted to written domain with a hand-crafted set of FST normalization rules.

For our purposes, it would be possible to train the E2E model to output hypotheses in
the spoken domain, and then to use either a neural network~\cite{Sproat17} or an
FST-based system~\cite{Vasserman15} to convert the hypotheses into the written
domain.
To keep overall system size as small as possible, we instead train the E2E model
to directly output hypotheses in the written domain (i.e., normalized into the
output form).
Since we do not observe a sufficiently large number of audio-text pairs
containing numeric sequences in training, we generate a set of 5 million
utterances containing numeric entities.
We synthesize this data using a concatenative TTS approach with one
voice~\cite{Gonzalvo16} to create audio-text pairs, which we augment to our
training data (each batch is filled with supervised data 90\% of the time and
synthetic data 10\% of the time).

\vspace{-3mm}
\section{Experimental Details}
\vspace{-3mm}
\label{sec:experiments}
\subsection{Data Sets}
The training set used for experiments consists of 35 million English utterances
($\sim 27,500$ hours).
The training utterances are anonymized and hand-transcribed, and are
representative of Google's voice search and dictation traffic.
This data set is created by artificially corrupting clean utterances using a
room simulator, adding varying degrees of noise and reverberation such that the
overall SNR is between 0dB and 30dB, with an average SNR of
12dB~\cite{Chanwoo17}.
The noise sources are from YouTube and daily life noisy environmental
recordings.
The main test sets we report results on include ∼14.8K voice search (\emph{VS})
utterances extracted from Google traffic, as well as 15.7K dictation utterances,
which we refer to as the \emph{IME} test set.

To evaluate the performance of contextual biasing, we report performance on 4
voice command test sets, namely \emph{Songs} (requests to play media),
\emph{Contacts-Real}, \emph{Contacts-TTS} (requests to call/text contacts), and
\emph{Apps} (requests to interact with an app).
All sets except \emph{Contacts-Real} are created by mining song, contact or app
names from the web, and synthesizing TTS utterances in each of these categories.
The \emph{Contacts-Real} set contains anonymized and hand-transcribed utterances
extracted from Google traffic.
Only utterances with an intent to communicate with a contact are included in the
test set.
Noise is then artificially added to the TTS data, similar to the process
described above~\cite{Chanwoo17}.

To evaluate the performance of numerics, we report results on a real data
numerics set (\emph{Num-Real}), which contains anonymized and hand-transcribed
utterances extracted from Google traffic. In addition, we include performance on
a synthesized numerics set (\emph{Num-TTS}), which uses Parallel Wavenet~\cite{Aaron17}
with 1 voice. No utterance / transcript from the numerics test set appears in
the TTS training set from Section \ref{sec:numerics}.

\subsection{Model Architecture Details}
All experiments use 80-dimensional log-Mel features, computed with a 25ms window
and shifted every 10ms.
Similar to~\cite{Golan16}, at the current frame, $t$, these features are stacked
with 3 frames to the left and downsampled to a 30ms frame rate.
The encoder network consists of 8 LSTM layers, where each layer has 2,048 hidden units
followed by a 640-dimensional projection layer.
For all models in this work, we insert a time-reduction layer with the reduction
factor $N=2$ after the second layer of encoder to achieve 1.7$\times$
improvement in overall system speed without any accuracy loss.
The prediction network is 2 LSTM layers with 2,048 hidden units and a
640-dimensional projection per layer.
The encoder and prediction network are fed to a joint-network that has 640
hidden units.
The joint network is fed to a softmax layer, with either 76 units (for
graphemes) or 4,096 units (for word-pieces~\cite{Schuster2012}).
The total size of the RNN-T model is 117M parameters for graphemes and 120M parameters for word-pieces. For the WPM, after quantization, the total size is 120MB.
All RNN-T models are trained in Tensorflow~\cite{AbadiAgarwalBarhamEtAl15} on $8
\times 8$ Tensor Processing Units (TPU) slices with a global batch size of
4,096.

In this work, we compare the RNN-T model to a strong baseline conventional CTC embedded model,
which is similar to~\cite{McGrawPrabhavalkarAlvarezArenasEtAl16} but much larger.
The acoustic model consists of a CI-phone CTC model with 6 LSTM layers, where
each layer has 1,200 hidden units followed by a 400-dimensional projection
layer, and a 42-phoneme output softmax layer.
The lexicon has 500K words in the vocabulary.
We use a 5-gram first-pass language model and a small and efficient second-pass
rescoring LSTM LM.
Overall the size of the model after quantization is 130MB, which is of similar
size to the RNN-T model.

\vspace{-3mm}
\section{Results}
 \label{sec:results}
\subsection{Quality Improvements}
Table~\ref{table:quality} outlines various improvements to the quality of
RNN-T models.
First, $E1$ shows that layer norm~\cite{Ba16} helps to stabilize training,
resulting in a 6\% relative improvement in WER for VS and IME.
Next, by moving RNN-T training to TPUs~\cite{Sim17} and having a larger batch
size, we can get between a 1--4\% relative improvement.
Finally, changing units from graphemes to
word-pieces~\cite{RaoSakPrabhavalkar17} ($E3$) shows a 9\% relative improvement.
Overall, our algorithmic changes show 27\% and 25\% relative improvement on VS
and IME respectively compared to the baseline conventional CTC embedded model
($B0$).
All experiments going forward in the paper will report results using layer norm,
word-pieces and TPU training ($E3$).
\begin{table}[h!]
\centerline{
    \begin{tabular}{|c|c|c|c|}
    \hline
    ID & Model & \emph{VS WER} & \emph{IME WER} \\ \hline
    E0 & RNN-T Grapheme & 8.1\% & 4.9\% \\ \hline
    E1 & +Layer Norm & 7.6\% & 4.6\% \\ \hline
    E2 & +Larger Batch & 7.5\% & 4.4\% \\ \hline
    E3 & +Word-piece & \textbf{6.8\%} & \textbf{4.0\%} \\ \hline \hline
    B0 & CTC & 9.3\% & 5.3\% \\ \hline
  \end{tabular}
}
\caption{\label{table:quality} RNN-T model improvements. All models are \emph{unquantized}.}
\end{table}

\vspace{-7mm}
\subsection{Contextual Biasing}
Table~\ref{table:biasing} shows results using the shallow-fusion biasing
mechanism. 
We report biasing results with just supervised data ($E4$) and also including
unsupervised data ($E6$).
We also show biasing performance for the CTC conventional model in $B1$.
The table indicates that E2E biasing outperforms or is on par with
conventional-model biasing on all sets, except songs likely because the
out-of-vocabulary rate in songs is 1.0\%, which is higher than contacts
(0.2\%) or apps (0.5\%).

\begin{table}[h!]
\centerline{
  \begin{tabular}{|p{0.4cm}|p{2.6cm}|p{0.7cm}|p{1.0cm}|p{1.0cm}|p{0.7cm}|}
    \hline
    ID & Model & \emph{Songs} & \emph{Contacts-Real} & \emph{Contacts-TTS} & \emph{Apps} \\ \hline
    E3 & RNN-T Word-piece & 20.0\%  & 15.9\% & 35.0\% & 13.1\%   \\ \hline
    E4 & + Biasing & 3.4\%  & 6.4\% & 7.1\% & 1.9\%   \\ \hline  \hline
    E5 & E3 + Unsupervised & 14.7\%  & 15.4\%  & 25.0\%  & 9.6\%   \\ \hline 
    E6 & + Biasing & 3.0\%  &  \textbf{5.8\%} &  \textbf{5.4\%}  & \textbf{1.7\%}   \\ \hline 
    \hline
    B1 & CTC + Biasing & \textbf{2.4\%} & 6.8\%  &  5.7\%  & 2.4\%  \\ \hline
  \end{tabular}
}
\caption{\label{table:biasing} WER on contextual biasing sets. All models \emph{unquantized}.}
\vspace{-6mm}
\end{table}

\vspace{-4mm}
\subsection{Text normalization}
Next, Table~\ref{table:numerics} indicates the performance of the baseline RNN-T
($E3$) word-piece model on two numeric sets.
As can be seen in the table, the WER on the \emph{Num-TTS} set is really
high.
A closer error analysis reveals that these are due to the text normalization
errors: e.g., if the user speaks \texttt{call two double three four ...}, the RNN-T
model hypothesizes \texttt{2 double 3 4} rather than \texttt{2334}.
To fix this, we train the RNN-T model with more numeric examples ($E7$), as
described in Section \ref{sec:numerics}, which mitigates this issue
substantially, at the cost of a small degradation on \emph{VS} and \emph{IME}.
However, we note that this still outperforms the baseline system with a separate
FST-based normalizer~\cite{McGrawPrabhavalkarAlvarezArenasEtAl16} ($B0$) on
all sets.
\begin{table}[h!]
\centerline{
  \begin{tabular}{|p{0.4cm}|p{2.5cm}|p{0.6cm}|p{0.6cm}|p{1.3cm}|p{1.3cm}|}
    \hline
    ID & Model & \emph{VS} & \emph{IME} & \emph{Num-Real} & \emph{Num-TTS} \\ \hline
    E3 & RNN-T Word-piece & 6.8\% & 4.0\% & 6.7\% & 22.8\% \\ \hline
    E7 & + numerics TTS & \textbf{7.0\%} & \textbf{4.1\%} & \textbf{6.9\%} & \textbf{4.3\%} \\ \hline \hline
    B0 & CTC & 9.3\% & 5.3\% & 8.4\% & 6.8\% \\ \hline
  \end{tabular}
}
\caption{\label{table:numerics} WER on numeric sets. All models are \emph{unquantized}.}
\end{table}

\vspace{-8mm}
\subsection{Real Time Factor \label{sec:rtf}}
In Table~\ref{table:quantization}, we report WER and RT90, i.e. real time factor
(processing time divided by audio duration) at 90 percentile, where lower
values indicate faster processing and lower user-perceived latency.
Comparing $E2$ and $E7$, we can see that the RNN-T word-piece model outperforms
the grapheme model in both accuracy and speed.

Quantization speeds up inference further: asymmetric quantization ($E8$)
improves RT90 by 28\% compared to the float model ($E7$) with only a 0.1\%
absolute WER degradation; symmetric quantization ($E9$), which assumes that weights
are centered around zero, only introduces additional small degradation on VS
WER, but leads to a substantial reduction in RT90 (64\% compared to the float
model), which is twice as fast as real time.
Moreover, quantization reduces model size by 4$\times$.
Our best model ($E9$) is also faster than the conventional CTC model $B2$, while
still achieving accuracy improvements of more than 20\%.
\begin{table}[h!]
\centerline{
  \begin{tabular}{|p{0.4cm}|p{4.0cm}|p{0.6cm}|p{0.6cm}|p{0.6cm}|}
    \hline
    ID & Model & \emph{VS} & \emph{IME} & \emph{RT90} \\ \hline
    E2 & RNN-T Grapheme (Float) & 7.5\% & 4.4\% & 1.58 \\ \hline
    E7 & RNN-T Word-piece (Float) & 7.0\% & 4.1\% & 1.43 \\ \hline
    E8 & + Asymmetric Quantization & 7.1\% & 4.2\% & 1.03 \\ \hline
    E9 & + Symmetric Quantization & \textbf{7.3\%} & \textbf{4.2\%} & \textbf{0.51} \\ \hline \hline
    B2 & CTC + Symmetric Quantization  & 9.2\% & 5.4\% & 0.86 \\ \hline
  \end{tabular}
}
\caption{\label{table:quantization} Quantization results on WER and RT90.}
\vspace{-0.1in}
\end{table}

\vspace{-6mm}
\section{Conclusions \label{sec:conclusions}}

We present the design of a compact E2E speech recognizer based on the RNN-T
model which runs twice as fast as real-time on a Google Pixel phone, and
improves WER by more than 20\% over a strong embedded baseline system on both
voice search and dictation tasks.
This is achieved through a series of modifications to the RNN-T model
architecture, quantized inference, and the use of TTS to synthesize
training data for the E2E model.
The proposed system shows that an end-to-end trained, all-neural model is
very well suited for on-device applications for its ability to perform
streaming, high-accuracy, low-latency, contextual speech recognition.

\footnotesize
\bibliographystyle{IEEEbib}
\bibliography{main}
\end{document}